%% file: main.tex
\documentclass[10pt,twocolumn,letterpaper]{article}

\input{packages}
\iccvfinalcopy % *** Uncomment this line for the final submission

\begin{document}

%%%%%%%%% TITLE
\title{\includegraphics[width=0.066\linewidth]{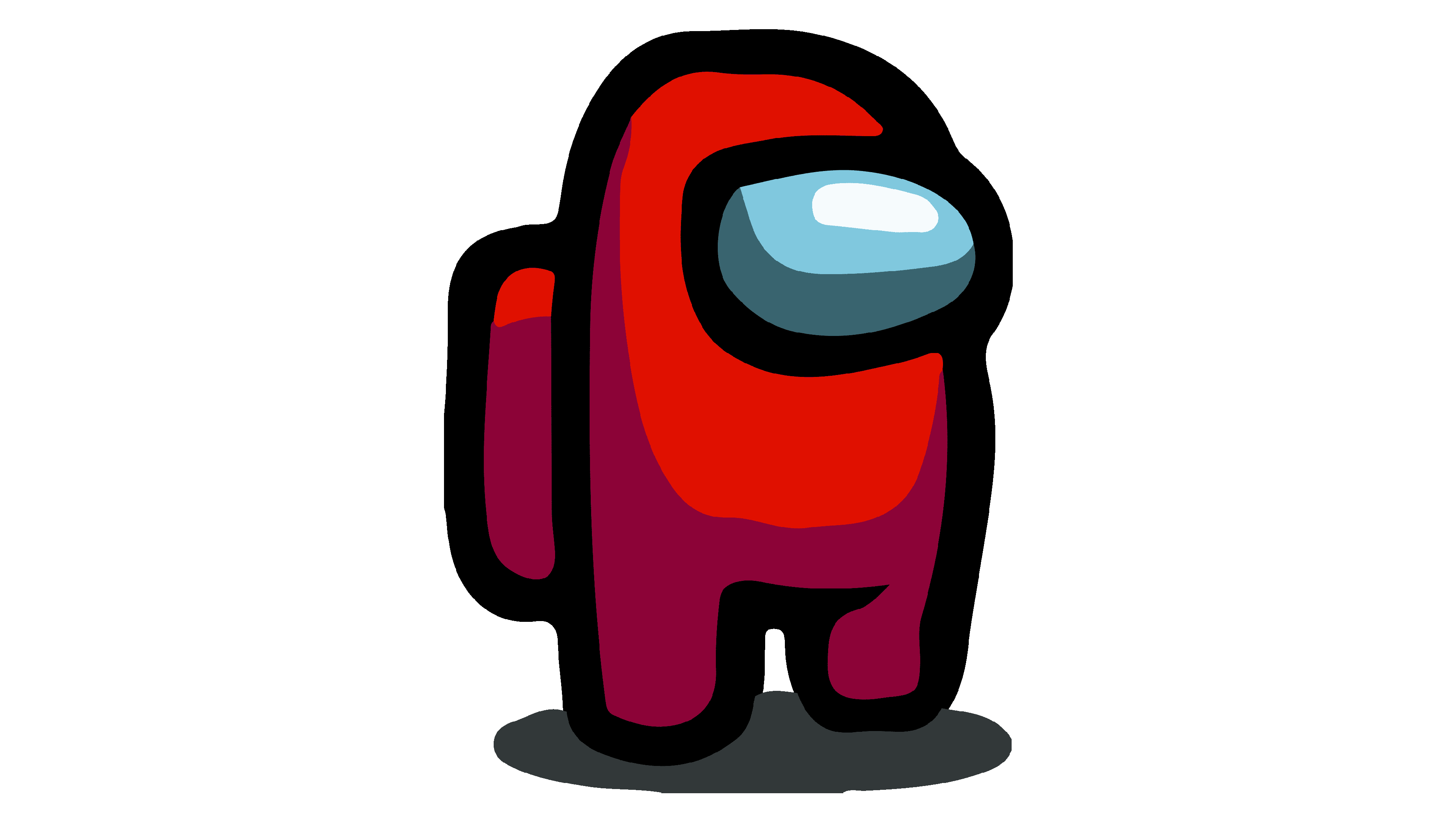}\textit{Among Us}: Adversarially Robust Collaborative Perception by Consensus}

\author{Yiming Li$^{1,*}$ \quad Qi Fang$^{1,}$\thanks{ 
 indicates equal contribution} \quad  Jiamu Bai$^1$ \quad Siheng Chen$^{2,3}$ \quad Felix Juefei-Xu$^4$ \quad Chen Feng$^{1,}$\thanks{Corresponding author. The work is supported by NSF grants 2238968 and 2026479.}\\
$^{1}$New York University \quad $^{2}$Shanghai Jiao Tong University \quad $^{3}$Shanghai AI Laboratory \quad $^{4}$Meta AI
\\
% {\tt\small yimingli@nyu.edu, cfeng@nyu.edu}
% {\tt\small \url{github}}
{\tt\small \{yimingli, qifang, jb7082\}@nyu.edu, sihengc@sjtu.edu.cn, felixu@meta.com, cfeng@nyu.edu} \\
{\tt\small \url{https://github.com/coperception/ROBOSAC}}
\vspace{-3mm}
}

\maketitle
\ificcvfinal\thispagestyle{empty}\fi

\pagestyle{empty}

\begin{abstract}
\vspace{-1mm}
% Multiple robots could perceive a scene (e.g., detect objects) collaboratively better than individually. Yet adversarial messages could easily fool such a collaboration that uses deep learning. 
Multiple robots could perceive a scene (e.g., detect objects) collaboratively better than individuals, although easily suffer from adversarial attacks when using deep learning. 
This could be addressed by the adversarial defense, but its training requires the often-unknown attacking mechanism.
Differently, we propose \acronym~, a novel sampling-based defense strategy generalizable to unseen attackers. Our key idea is that collaborative perception should lead to consensus rather than dissensus in results compared to individual perception. This leads to our hypothesize-and-verify framework: perception results with and without collaboration from a random subset of teammates are compared until reaching a consensus. 
In such a framework, more teammates in the sampled subset often entail better perception performance but require longer sampling time to reject potential attackers. Thus, we derive how many sampling trials are needed to ensure the desired size of an attacker-free subset, or equivalently, the maximum size of such a subset that we can successfully sample within a given number of trials. We validate our method on the task of collaborative 3D object detection in autonomous driving scenarios.
% Since better performance often needs more teammates, which then requires longer computation to reject attackers, we derive the maximum size of the teammate subset at a given sampling budget, and also the maximum sampling budget to ensure a desired number of attacker-free teammates, all for achieving a guaranteed consensus probability. 
\vspace{-2.5mm}
\end{abstract}

\section{Introduction}
	Perception is a fundamental capability for autonomous robots to understand their surroundings~\cite{ma2022vision,yin2021modeling,li2020lidar}. Single-robot perception suffers from long-range or occlusion issues which stem from limited sensing capabilities and inadequate individual viewpoints~\cite{roldao20223d}. Therefore, collaborative perception (co-perception) is proposed to provide more viewpoints for each robot via communication, so that robots can see further and better~\cite{wang2020v2vnet,li2022multi,xu2022cobevt}.
 
 In literature, raw-data-level and decision-level fusion both demonstrate satisfactory performance in terms of robustness and precision~\cite{Chen2019CooperCP, arnold2020cooperative}. The recent development of deep learning has revolutionized many fields including robotic perception, and feature-level fusion has been proposed in which intermediate representations from a deep neural network (DNN) are shared amongst robots. Unlike raw-data-level and decision-level fusion approaches, feature-level fusion presents the advantages of good compressibility and the preservation of contextual information, further enhancing the performance-bandwidth trade-off in multi-robot perception~\cite{liu2020when2com,wang2020v2vnet,li2021learning,su2022uncertainty}.

\begin{figure}
    \centering
    \includegraphics[width=\linewidth]{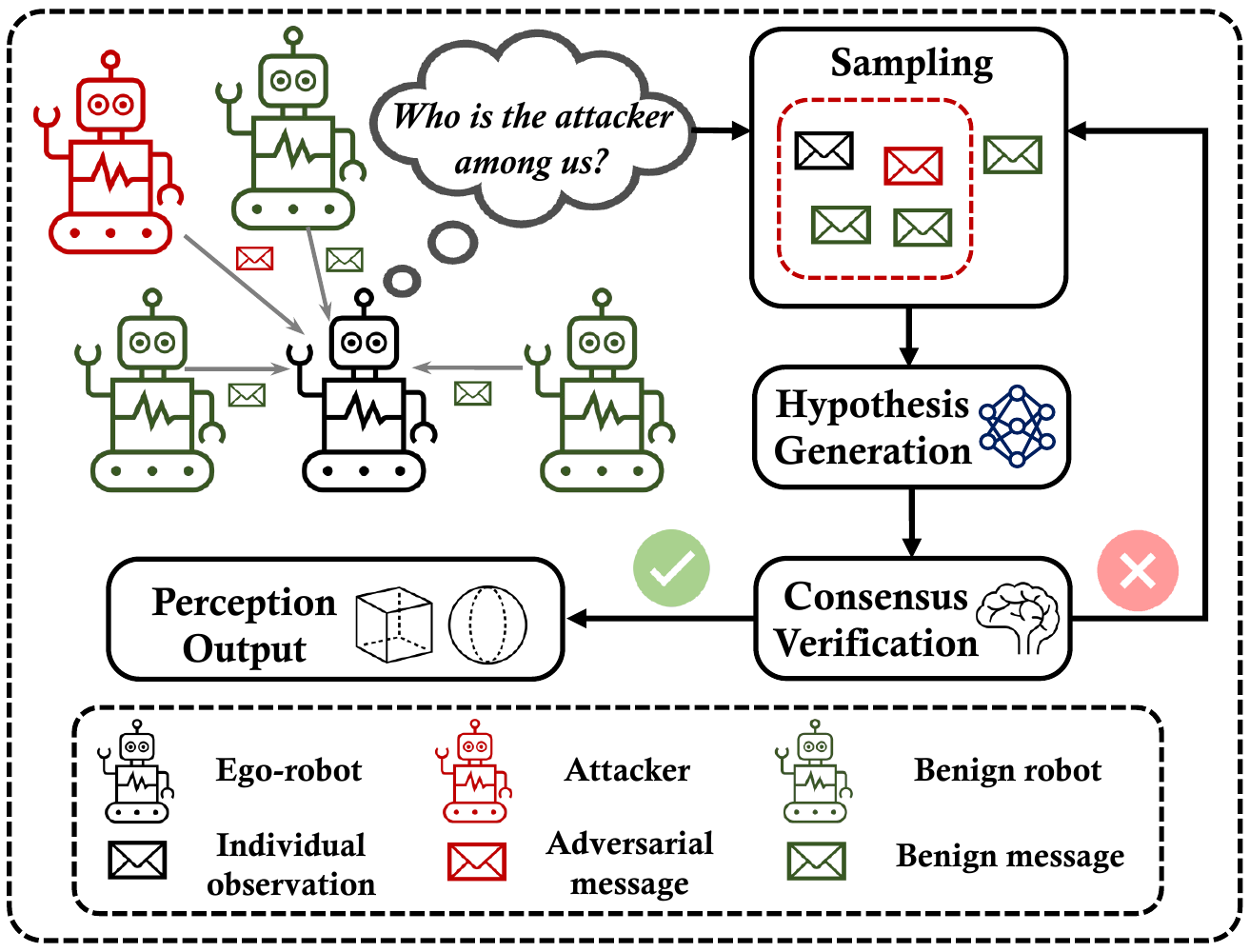}
    \vspace{-5mm}
    \caption{\textbf{Overview of \acronym~.} Ego-robot aims to find several benign collaborators with a hypothesize-and-verify procedure until reaching a consensus or using up the sampling budget. Consensus is checked between the results with and without the selected teammates.}
    \label{fig:teaser}
    \vspace{-2.5mm}
\end{figure}

	Although the original motivation for collaborative perception is to promote resilience and robustness via information sharing, the communication channel could potentially become a wide-open backdoor in DNN-based perception models due to the well-known adversarial vulnerability of DNNs~\cite{Szegedy2014IntriguingPO}. Prior work has shown that a maliciously-crafted imperceptible perturbation added on the shared feature can drastically alter the perception output, jeopardizing the perception system~\cite{tu2021adversarial}. To solve the safety concerns, adversarial training has been exploited~\cite{tu2021adversarial}, yet it introduces extra overhead during training and fails to generalize to unseen attackers~\cite{zhang2019adversarial}. Besides, adversarial training may lead to a small loss of accuracy~\cite{raghunathan2020understanding}. In a word, it is still non-trivial to achieve  \textit{computationally-efficient} and \textit{generalizable} adversarial defense in collaborative perception.
	
    In this work, different from applying adversarial training after indiscriminately using all messages, we propose to enable the ego-robot to intelligently select benign collaborators from teammates, instead of naively trusting all the teammates. 
    % On one hand, the ego-robot cannot totally trust others and need to carefully use shared messages to avoid being attacked. On the other hand, it cannot fully ignore the messages to keep away from attackers, since they still need complementary information for better perception. 
    % We consider that a team of robots is perceiving the 3D environment, and due to the limited field of view for an individual, each robot tries to take advantage of the messages shared by others to understand the environment comprehensively. However, malicious robots make up a portion of the team, and they send out adversarial messages to fool the others' perception systems. Consequently, on one hand, each robot cannot totally trust the others and need to carefully use the shared messages to avoid being attacked. On the other hand, they also cannot fully ignore the messages from others to keep away from attackers, since they still need complementary information for better perception.
    Inspired by random sample consensus (RANSAC) in robust estimation~\cite{martinez2022ransac}, we propose \textbf{ROB}ust c\textbf{O}llaborative \textbf{SA}mple \textbf{C}onsensus (\acronym~),
    % multi-\textbf{ROB}ot c\textbf{O}llaboration \textbf{SA}mple \textbf{C}onsensus (\acronym~),
    a general sampling-based framework for adversarially robust collaborative perception. Our key idea is that the robot is supposed to reach a consensus with its teammates after collaboration, rather than largely diverging from its individual perception. Specifically, \acronym~ utilizes the hypothesize-and-verify workflow: the robot samples a subset of teammates and compares the results with and without the sampled teammates. After the consensus is verified, indicating no attackers \textit{among us}, the robot can output the perceptual results generated in collaboration with teammates for further decision-making, as shown in Fig.~\ref{fig:teaser}. Different from the widely-used adversarial training, \acronym~ is attacker-agnostic and thus can easily generalize to unseen adversarial learning algorithms. 
    
    % \Note{sc: emphasize the advantages of this method, for example, computationally-efficiency and can be generalized to unseen attackers, which addresses the bottleneck issues in previous adverbial training.}
    
    Meanwhile, \acronym~ can be customized for either strong performance or high efficiency: as more benign teammates leading to better performance require more computation to reject the attackers, there exists a performance-computation trade-off in \acronym~. Formally, under various attacker ratios, we can compute the \textit{maximum number of attacker-free collaborators} that could be found given a specific sampling budget, and the \textit{upper bound of the number of sampling} ensuring a desired number of benign teammates, all for achieving a guaranteed consensus probability. Additionally, we propose an adaptive probing approach to handle the scenario of unknown attacker ratios, starting from trusting all the teammates and then gradually becoming more vigilant. Our contributions are summarized as:
    \begin{itemize}[leftmargin=1.3em]
\item We develop \acronym~, a scalable, generalizable, and generally-applicable adversarially robust collaborative perception framework via multi-robot consensus. 

\item We propose aggressive-to-conservative probing (A2CP) with retrospect to estimate the attacker ratio efficiently.

% \Note{sc: i cannot see this in the intro and abstract.}

\item We conduct experiments on collaborative 3D object detection in safety-critical autonomous driving to validate the effectiveness of \acronym~.
\end{itemize}

%===============================================================================

\section{Related Works}\label{sec:related_works}
\textbf{Collaborative perception.}
To solve the fundamental issues of single-robot perception such as limited field-of-view~\cite{cao2022monoscene, 10015799, li2023voxformer,  10015867}, multi-robot collaboration has been exploited to ameliorate the precision, robustness, and resilience of the perception system~\cite{li2022v2x}. Previous works primarily investigate multi-robot perception in the aerial scenario~\cite{liu2020when2com, liu2020who2com, zhou2022multi} and autonomous driving~\cite{Chen2019CooperCP, su2023collaborative, Kim2015MultivehicleCD}, in different tasks such as object detection, semantic segmentation, and depth estimation. There are three kinds of communication strategies in multi-robot perception: (1) raw-data-level fusion, (2) feature-level fusion, and (3) output-level fusion. Amongst all three methods, feature-level fusion transmits learned intermediate representations of deep neural networks. Since the intermediate representations are easy to compress and are equipped with contextual knowledge of the environment, feature-level fusion demonstrates better performance-bandwidth trade-off, thus is widely applied in autonomous robots~\cite{liu2020when2com, wang2020v2vnet, li2021learning, vadivelu2020learning}. Nevertheless, the adversarial robustness of feature-level fusion is underexplored.

% \Note{sc: what is the connection to this work?}

\textbf{Adversarial perception.} Adversarial vulnerability in DNNs~\cite{Szegedy2014IntriguingPO} can endanger the learning-based single-robot perception systems in safety-critical scenarios like autonomous driving~\cite{Tu2020PhysicallyRA,Li_2021_ICCV,Cao2019AdversarialSA,tu2021exploring}. For multi-robot perception, \cite{tu2021adversarial} reveals that an indistinguishable adversarial noise added on the shared intermediate representation can result in a lot of false detections. Adversarial attack can be classified into white-box attack and black-box attacks~\cite {ren2020adversarial}. White-box attacker has full information about the DNNs~\cite{Szegedy2014IntriguingPO,goodfellow2014explaining,kurakin2018adversarial}, while black-box attacker is usually less effective than white-box attack since attackers have no access to the target models. The information about the model is obtained through query~\cite{brendel2018decisionbased} or inferred through highly transferable surrogate models~\cite{cheng2019improving,xie2019improving}. For defense, adversarial training is proposed by incorporating adversarial samples into training stages, yet it requires the knowledge of attackers. Hence, to realize a generalizable adversarial defense against unseen attackers is still non-trivial and underexplored. 

% Another kind of defense strategy is carried out during inference, such as randomization (treating the adversarial attacks as random effects)~\cite{xie2018mitigating,pinot2020randomization} and denoising~\cite{xie2019feature}. 

% \Note{sc: what is the connection to this work?}

% (applying self-supervised learning technique to revert the input samples back to their original values). 

\textbf{RANSAC.}
Random sample consensus (RANSAC) is a well-known robust estimation algorithm applicable to datasets containing a number of outliers. It was first proposed by Fischler and Bolles to solve Location Determination Problem (LDP)~\cite{fischler1981random}. RANSAC employs a hypothesize-and-verify pipeline to iteratively select a sample of data points in the process of fitting the optimal model. At first, RANSAC was largely applied in works in the image domain, and gradually extension work of RANSAC is proposed, \textit{e.g.}, MSAC (M-estimator SAmple and Consensus)~\cite{torr1998robust} and MLESAC (Maximum Likelihood Estimation SAmple and Consensus)~\cite{torr2000mlesac}. Currently, RANSAC is widely used in computer vision~\cite{hodan2020epos,valassakis2022learning} and robotics~\cite{tanaka2006incremental}, \textit{e.g.}, to estimate the fundamental matrix and remove outlier correspondences in Structure from Motion (SfM)~\cite{ullman1979interpretation}.  In this work, we for the first time exploit the idea of sample consensus in the problem of robust collaborative perception.

\section{\acronym~}
\label{sec:general_ransac}
In this section, we present the problem setup for collaborative perception under adversarial attacks in~\cref{subsec:setup}, followed by the revisiting of  RANSAC in~\cref{subsec:ransac} and the illustration of a general defense framework termed \acronym~ in~\cref{subsec:mongsac1} (attacker ratio known) and~\cref{subsec:mongsac2} (attacker ratio unknown).

% Considering that a team of robots is perceiving the 3D environment, and due to the limited field of view, each robot tries to take advantage of the messages shared by others to achieve a comprehensive understanding of the environment. However, since there are malicious robots within the team, the key problem becomes how to enable each \textit{ego-robot} to make the most of information supplied by others while avoiding being attacked.

\subsection{Problem setup}\label{subsec:setup}
\textbf{Terminology.} We first introduce our terminology in this work. Benign robots that share their truly-observed information are termed \textit{collaborators}. Adversarial robots that share carefully-crafted harmful messages are termed \textit{attackers}. The robot that tries to exploit the collaborators' information while  protecting itself from attackers is termed \textit{ego-robot}. All the robots other than the \textit{ego-robot} are termed \textit{teammates}, including both \textit{collaborators} and \textit{attackers}.

\textbf{Assumption and setup.} We consider two scenarios: (1) \textit{static team}: the ego-robot communicates with the same teammates, and (2) \textit{dynamic team}: the ego-robot meets and communicates with different teammates.
We assume that the attacker ratio is fixed in both scenarios. 
% Regarding scenario (1), it is reasonable and realistic to keep the role of each robot unchanged within a period of time. Regarding scenario (2), it is more challenging than (1) because the robot can alter its intention, \textit{e.g.}, a benign robot is suddenly hacked or a hacked robot is recovered. 
\textit{Regarding the attack}, attackers have access to the ego-robot's viewpoint and utilize adversarial learning to generate imperceptible noises added to its original messages, to significantly degrade the output of the ego-robot. \textit{Regarding the defense}, the ego-robot is not aware of the specific attacking strategy, but it can identify whether it has been attacked based on the change in output space.

\textbf{Objective and challenge.} 
% We assume that a team of robots is perceiving the 3D environment, and due to limited field of view for an individual, each robot tries to take advantage of the messages shared by others to achieve a comprehensive understanding of the environment. However, malicious robots make up a portion of the team, and they send out adversarial messages in order to attack the benign robots' perception systems. 
On one hand, the ego-robot cannot totally trust others and need to carefully use the messages to avoid being attacked. On the other hand, they also cannot fully ignore the messages to keep away from attackers, since they still need complementary information to enhance their own restricted perception. The objective for each ego-robot is: \textit{given a certain computation budget for consensus verification, how to make full use of the messages shared by others while avoiding being attacked?} The challenges for this problem lie in two aspects: (1) \textit{generalizability:} how to achieve generalizable defense given that there could be different and unseen attackers; and (2) \textit{scalability:} how to realize   computationally-efficient defense especially when there are a large number of teammates.

% \Note{sc: do we need to mathematically formulate the problem here? for example, be explicit that with probability $\eta$, a robot would be an attacker.}

% each robot can change its mind about whether to attack or not, thus the ego-robot fails to rely on its past judgements on its teammates; (2) attackers must be identified as quickly as possible to prevent extra latency in the ego-robot's decision making. 

% Original Algorithm Part

\begin{algorithm}[h]
\small
	%\textsl{}\setstretch{1.8}
	\renewcommand{\algorithmicrequire}{\textbf{Input:}}
	\renewcommand{\algorithmicensure}{\textbf{Output:}}
	\newcommand{\algorithmicbreak}{\textbf{break}}
        \newcommand{\BREAK}{\STATE \algorithmicbreak}
        
	\caption{Workflow of \acronym~}
	\label{alg:ransac}
        \begin{adjustwidth}{0pt}{12pt}
	\begin{algorithmic}[1]
	    \REQUIRE The total number of teammates $S$, messages from teammates $\{ \mathbf{M}_i \}_{i=1,\ldots,S}$, message of an ego-robot $\mathbf{M}_0$, perception model $f_\theta$, difference measure $d$, consensus threshold $\epsilon$, sampling budget $N$, attacker ratio $\eta$, probability of at least one successful sampling $p$ 
	    \ENSURE The perception results for the ego-robot $\mathbf{Y}_0$ at current timestamp
	    \STATE  Obtain the perception results of only using the ego-robot's message: $\hat{\mathbf{Y}}_0 = f_\theta(\mathbf{M}_0)$
		\STATE Calculate the guaranteed maximum number of collaborators :$s = \lfloor \frac{\ln{[1 -(1-p)^{\frac{1}{N}}]}}{\ln{(1-\eta)}} \rfloor$
		\STATE $n = 0$
		\WHILE{$n<N$}
		\STATE $n \leftarrow n + 1$
		% \STATE Sample $s$ teammates $\{ \mathbf{M}_j \}_{j=1,\ldots,s}$ \hfill $\vartriangleright$ \texttt{Random sample}
            \STATE Sample $s$ teammates randomly $\{ \mathbf{M}_j \}_{j=1,\ldots,s}$ 
		\STATE Obtain the perception results $\hat{\mathbf{Y}}_s = f_\theta(\mathbf{M}_0, \{ \mathbf{M}_j \}_{j=1,\ldots,s})$
		\IF[$\vartriangleright$ \texttt{Consensus}]{$ d(\hat{\mathbf{Y}}_s, \hat{\mathbf{Y}}_0) \leq \epsilon$}  
		\STATE $\mathbf{Y}_0 =\hat{\mathbf{Y}}_s$ 
% 		\hfill $\vartriangleright$ \texttt{Consensus verification}
		\BREAK \hfill $\vartriangleright$  \texttt{Early stop}
 		\ELSIF[\hfill $\vartriangleright$  \texttt{Dissensus}]{ $n = N-1$}
%  		\ELSIF{$d(\hat{\mathbf{Y}}_s, \hat{\mathbf{Y}}_0) > \epsilon$ \AND $n = N-1$}
		\STATE $\mathbf{Y}_0 =\hat{\mathbf{Y}}_0$ \hfill $\vartriangleright$  \texttt{No collaboration}
		\BREAK
		\ENDIF 
		\ENDWHILE
% 		\UNTIL $ < \varepsilon $  
% 		\ENSURE  decomposed modes $ \left\{ {{s_k}\left( t \right)} \right\}$, $\left\{ {{\omega _k}\left( t \right)} \right\}$
	\end{algorithmic} 
        \end{adjustwidth}
        % \hspace{-20pt}
\end{algorithm}

\subsection{Revisit of RANSAC}\label{subsec:ransac}
% We begin by introducing the classic RANSAC algorithm for robust estimation as a preface to presenting our method.
RANSAC uses a hypothesize-and-verify workflow to robustly fit a model based on a set of data points in the presence of noise. The workflow mainly includes three steps: (1) produce a set of model hypotheses by sampling minimal data points for the fitted model; (2) evaluate hypotheses with some consensus metrics such as the number of inliers; (3) choose the best hypothesis. An optional fourth step is often employed in practice, which refines the chosen hypothesis with all inliers. RANSAC can compute the required number of sampling to ensure a high probability of at least one successful sampling under different outlier ratios. 

% \Note{sc: overlap with related works?}

% \subsection{Our method}
% To solve the above problem, one naive solution is to iteratively verify each teammate based on the consensus in the output space. Take the 3D object detection task as an example, if the output bounding boxes are significantly different from the results given by the ego-robot alone after fusing someone's message, then the corresponding teammate could be identified as an attacker. However, such kind of method requires checking every teammate, thus is not scalable. We instead propose to iteratively sample a subset of teammates until they reach a consensus. 

% In the following, we will respectively introduce the original RANSAC algorithm and our adaptation, as well as the performance-computation trade-off in our case.

\subsection{Workflow of \acronym~}\label{subsec:mongsac1} 
One naive solution to the problem is iteratively verifying each teammate based on the consensus in the output space, yet it is not scalable. We instead propose to iteratively sample a subset of teammates until they reach a consensus, given a certain sampling budget. Assume that the attacker ratio is known as $\eta$, and the ego-robot plans to use the information from $s$ teammates so that it keeps on sampling $s$ teammates until reaching a consensus. The sampling budget is $N$, and a successful sampling is one that contains no attackers amongst the sampled $s$ robots. The detailed procedures for an ego-robot are: (1) produce perception results using its individual observation; (2) sample $s$ teammates and fuse their messages to generate perception results; (3) verify the consensus between the results in (1) and (2); (4) output the results in (2) if there are no attackers, otherwise, continue to implement (2). Formally, our objective is to maximize the probability of at least one successful sampling which is calculated by $p = 1-[1-(1-\eta)^s]^N$. The workflow is shown in Algorithm~\ref{alg:ransac}.

\begin{figure}[t]
\centering
   \vspace{-2mm}
\subfigure[\small $s$ v.s. $N$ ]{
\begin{minipage}[t]{0.495\linewidth}
\centering
\includegraphics[width=1\textwidth]{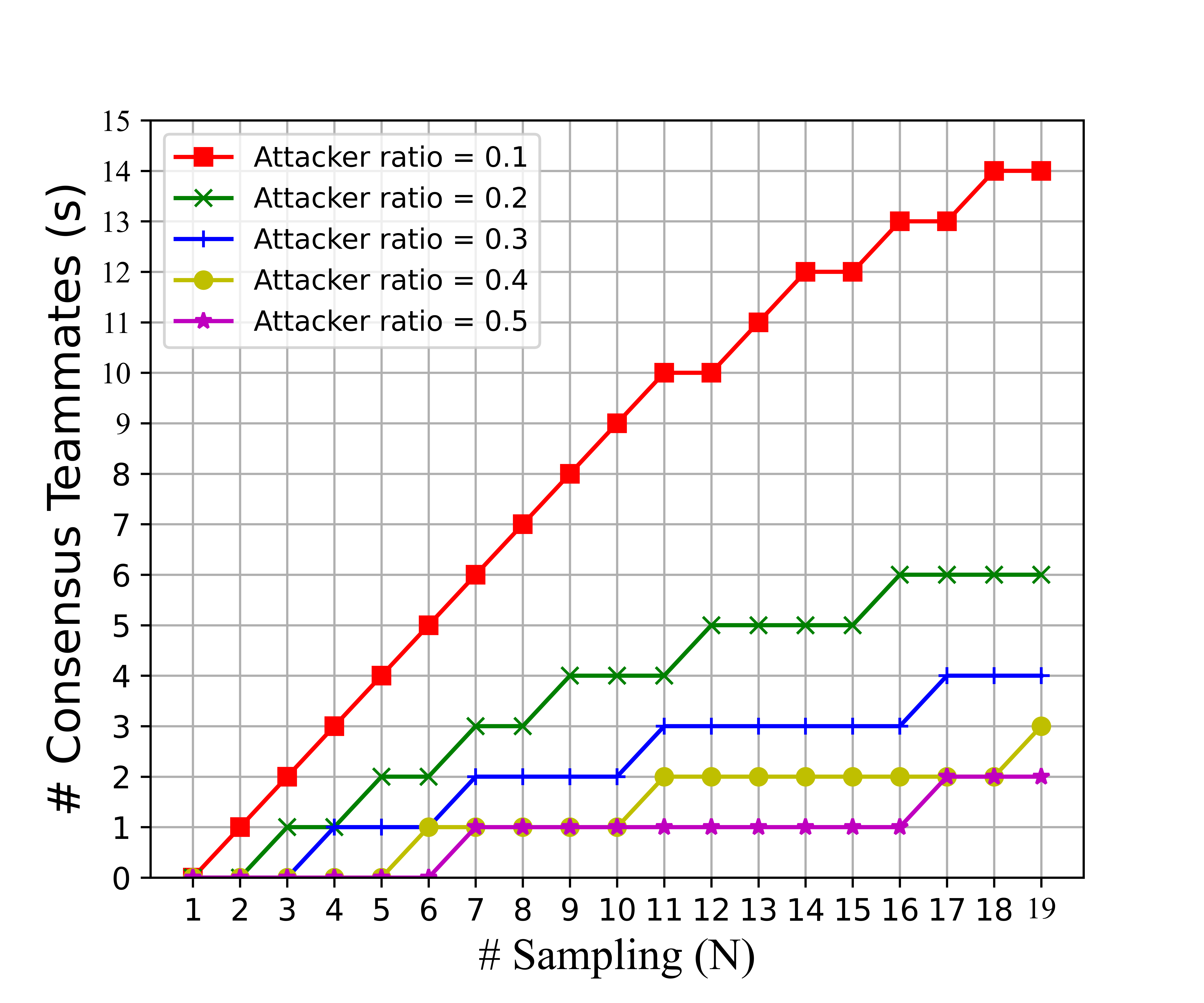}
%\caption{fig1}
\end{minipage}%
}%
\subfigure[\small $N$ v.s. $s$]{
\begin{minipage}[t]{0.495\linewidth}
\centering
\includegraphics[width=1\textwidth]{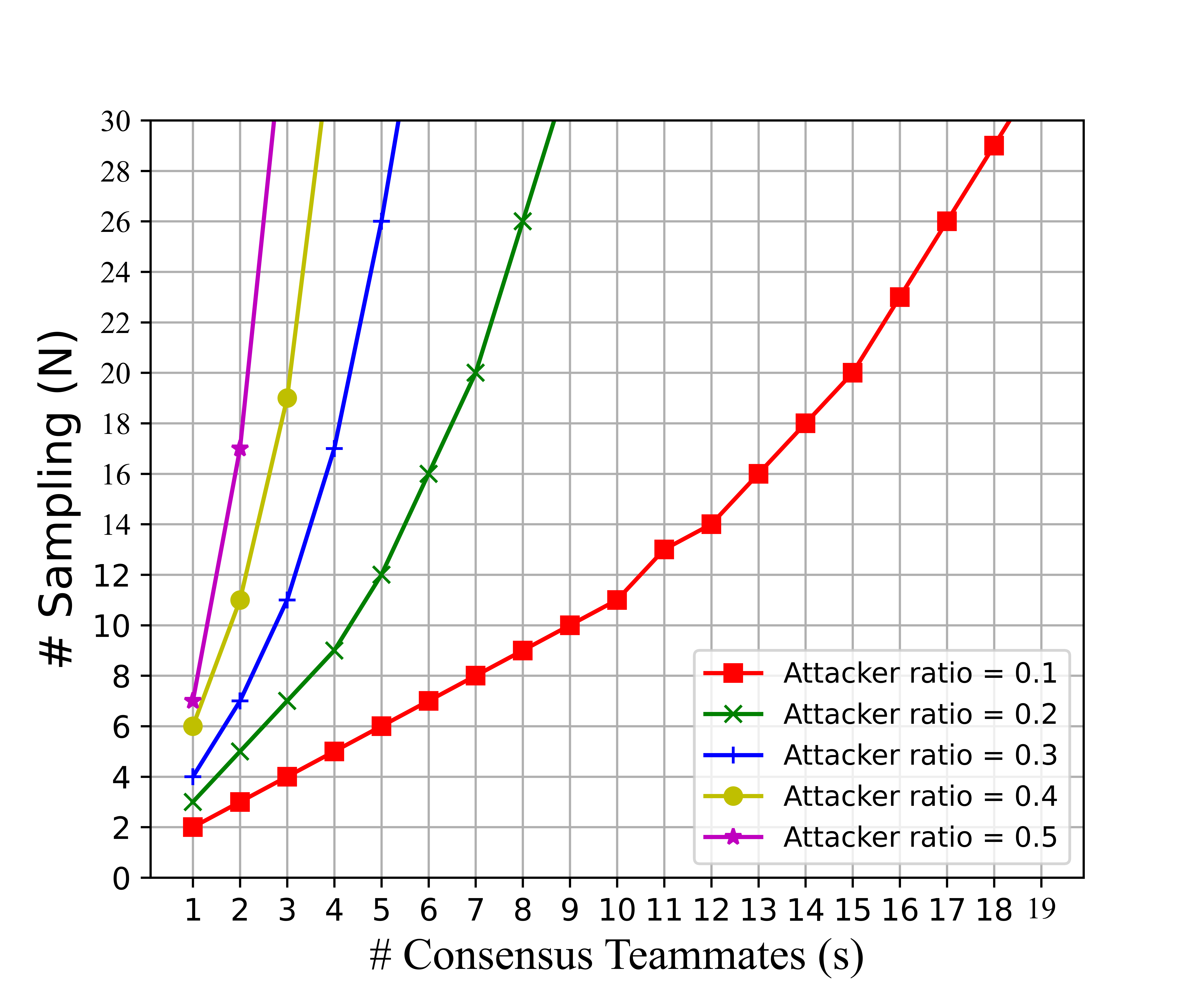}
% \caption{fig2}
\end{minipage}%
}%
\centering
	\vspace{-2mm}
\caption{\textbf{A numerical example of \acronym~} (probability of at least one successful sampling is $0.99$). (a) Guaranteed maximum number of collaborators given a certain sampling budget. (b) The maximum number of sampling given a desired number of collaborators.}
\label{fig:numerical}
\vspace{-5mm}
\end{figure}

\textbf{Performance-computation trade-off.}
Different from RANSAC, \acronym~ can be customized for the computation (budget $N$) or the performance related to the amount of beneficial information (number of collaborators $s$). Given a probability $p$ to ensure that there is at least one successful sampling within a sampling budget $N$, the maximum number of attacker-free collaborators that could be found is:
\begin{equation}\label{eq:sampling_size}
    s = \lfloor \frac{\ln{[1 -(1-p)^{\frac{1}{N}}]}}{\ln{(1-\eta)}} \rfloor,
\end{equation}
which means that the ego-robot is able to have $s$ collaborators to enhance its perception in a safe manner. In turn, given a probability $p$ to ensure that there is at least one successful sampling and a desired number of collaborators $s$, we need to sample at most:
\begin{equation}\label{eq:sampling_number}
    N = \lceil \frac{\ln{(1-p)}}{\ln{[1 - (1-\eta)^{s}]}} \rceil,
\end{equation}
which means that the sampled teammates can reach a consensus before the $N$-th sampling. One numerical example is illustrated in Fig.~\ref{fig:numerical}. Here the probability of at least one successful sampling is 0.99. In (a), given a sampling budget $5$, the ego-robot can make sure to have $4$ collaborators to improve the perception when the attacker ratio is $10\%$; in (b), given a desired number of collaborators $5$, the ego-robot can make final decisions before the $12$-th sampling when the attacker ratio is $20\%$.

\begin{algorithm}[t]
\small
%\textsl{}\setstretch{1.8}
\renewcommand{\algorithmicrequire}{\textbf{Input:}}
\renewcommand{\algorithmicensure}{\textbf{Output:}}
\newcommand{\algorithmicbreak}{\textbf{break}}
    \newcommand{\BREAK}{\STATE \algorithmicbreak}
\caption{Workflow of A2CP with Retrospect}
\label{alg:probing}
\begin{adjustwidth}{0pt}{12pt}
\begin{algorithmic}[1]
    \REQUIRE The number of teammates $S$, messages from teammates $\{ \mathbf{M}_i \}_{i=1,\ldots,S}$, message of an ego-robot $\mathbf{M}_0$, perception model $f_\theta$, difference measure $d$, consensus threshold $\epsilon$, ascending-ordered array of discretized possible ratios $\{ {\mathbf{R}}_k = {h_k}/S \}_{k= 1,2,\ldots,K}$ ($h_k \in [1..S)$), sampling budget $N > K$, probability of at least one successful sampling $p$ 
    \ENSURE The estimated attacker ratio $\hat{\eta}$
    \STATE $ \hat{\eta} \leftarrow 1.0$ 
    \STATE $\{{\mathbf{U}}_{k}\}_{k=1,\ldots,K } \leftarrow \mathbf{0}$ \hfill $\vartriangleright$ \texttt{Upper-bound of attempts}
    \STATE $\{{\mathbf{T}}_{k}\}_{k=1,\ldots, K} \leftarrow \mathbf{0}$ \hfill $\vartriangleright$ \texttt{Counter of attempts}
    \FOR{$k$ in $[1,\dots,K]$}
        % \STATE Calculate the maximum steps to ensure a consensus of given ratio $\eta$ and corresponding benign $S*(1-\eta)$ agents
        \STATE $\mathbf{U}_k = \lceil \frac{\ln{(1-p)}}{\ln{[1 - (1-\mathbf{R}_k)^{S(1-\mathbf{R}_k)}]}} \rceil$
        % \STATE Save each maximum step into array $\{U\}$
    \ENDFOR
    \FOR{each frame in the scene}
        % \IF{$\mathbf{T} < \mathbf{U}$}
            \STATE $n = 0$
            \STATE  Obtain the individual results: $\hat{\mathbf{Y}}_0 = f_\theta(\mathbf{M}_0)$
            \WHILE{$n<N$ and $\mathbf{T} < \mathbf{U}$}
                % \FOR{ratios $\eta$ in $\{ \mathbf{A} \}$}
                \FOR{$k \in \left[1,\ldots, K\right]$}
                    \IF{$\mathbf{T}_{k}$ $<$ $\mathbf{U}_{k}$}
                        \STATE $n \leftarrow n + 1$
                        \STATE $\eta \leftarrow \mathbf{R}_{k}$
                        \STATE Sample $S(1-\eta)$ teammates $\{ \mathbf{M}_j \}_{j=1,\ldots,S(1-\eta)}$ 
                        % \hfill $\vartriangleright$ \texttt{Random sample teammates based on the assumed attacker ratio}
                        % \STATE Obtain verification result $\mathbf{R}$ from \textbf{Algorithm \ref{alg:consensus_verification}}
                        \STATE Obtain the perception results $\hat{\mathbf{Y}}_s = f_\theta(\mathbf{M}_0, \{ \mathbf{M}_j \}_{j=1,\ldots,S(1-\eta)})$
                        \IF[$\vartriangleright$ \texttt{Consensus}]{$ d(\hat{\mathbf{Y}}_s, \hat{\mathbf{Y}}_0) \leq \epsilon$}
                            \STATE $\hat{\eta} \leftarrow \eta$ \hfill $\vartriangleright$ \texttt{Update}
                            \STATE { $\mathbf{T}_k  \leftarrow \mathbf{U}_k, ..., \mathbf{T}_K  \leftarrow \mathbf{U}_K$  \\ $\vartriangleright$  \texttt{Stop probing higher ratios}}
                            % \STATE {\scriptsize $\{\mathbf{T}_k, \ldots, \mathbf{T}_K \} \leftarrow \{\mathbf{U}_k, \ldots, \mathbf{U}_K \}$  $\vartriangleright$ \texttt{Stop to retrospect}}
                            % \STATE $\mathbf{Y}_0 =\hat{\mathbf{Y}}$ 
                            \BREAK
                        \ELSE[$\vartriangleright$ \texttt{Dissensus}]
                            \STATE $\mathbf{T}_{k}  \leftarrow \mathbf{T}_{k} + 1$ 
                            \ENDIF
                        \ENDIF
                \ENDFOR 
            \ENDWHILE
        % \ELSE
            % \BREAK
        % \ENDIF
    \ENDFOR
            % \WHILE{$\mathbf{T} < \mathbf{U}$}
            
  % \ENDWHILE
% 		\UNTIL $ < \varepsilon $  
% 		\ENSURE  decomposed modes $ \left\{ {{s_k}\left( t \right)} \right\}$, $\left\{ {{\omega _k}\left( t \right)} \right\}$
	\end{algorithmic}  
\end{adjustwidth}
\end{algorithm}

\subsection{Attacker Ratio Estimation}\label{subsec:mongsac2} 
The ego-robot might not be aware of the attacker ratio when it enters a novel environment. To estimate the attacker ratio as quickly as possible, we develop an \textbf{A}ggressive-to-\textbf{C}onservative \textbf{P}robing (A2CP) approach, which starts from trusting all teammates and gradually reduces the number of sampled teammates if the previous attempt fails. The basic idea is that once we find a consensus subset with $s$ robots, the ratio of benign collaborators will not be less than $\frac{s}{S}$, where $S$ is the total number of teammates. Take $S = 5$ as an example, we pre-define an array of discretized possible ratios in ascending order $ \mathbf{R} = [0.0, 0.2, 0.4, 0.6, 0.8]$ and start from probing $\eta=0.0$ ($s=5$) which indicates no attackers. If the first attempt succeeds, there is no need to probe other ratios and the ego-robot can collaborate with all teammates. If not, we continue to probe a higher ratio with fewer collaborators (in this case $\eta=0.2$ and $s=4$) and stop testing the unprobed ratios once consensus is verified. 

Note that we further propose a retrospect-based mechanism to avoid missing the probed lower ratios due to randomness. For example, if probing of $\eta=0.2, s=4$  fails yet probing of $\eta=0.4, s=3$ succeeds, we cannot conclude the ratio yet because it is more difficult for the previous more aggressive attempt to reach a consensus. Actually, there could be either 1 or 2 attackers. We continue to probe $\eta=0.2, s=4$ until enough attempts have been made. We use $\mathbf{T}_{k}$ to count the number of probing for a ratio $\mathbf{R}_{k}$, and $\mathbf{U}_{k}$ to denote the upper bound probing attempt of this ratio, which is derived from equation~\eqref{eq:sampling_number}. See Algorithm~\ref{alg:probing} for the overall workflow of the proposed attacker ratio estimator.

\section{Adversarially Robust Collaborative 3D Detection with \acronym~}
% \Note{sc: emphasize the relation to the previous section.}
In this section, we apply our \acronym~ framework in collaborative 3D object detection under adversarial attacks in autonomous driving. Vehicle-to-vehicle (V2V) communication could improve the robustness, safety, and efficiency of autonomous driving systems in light of more viewpoints and computational resources. However, there are still safety concerns regarding the communication channel~\cite{tu2021adversarial}. We consider that multiple vehicles located in the same geographical location are sharing information. An \textit{ego-vehicle} attempts to maximize the usage of benign \textit{collaborators} while defending against \textit{attackers}.

\textbf{Preliminaries.}
Each vehicle indexed by $i$ $(i=0,\ldots,S)$ is equipped with a 3D LiDAR sensor to generate a bird's eye view (BEV) occupancy grid map $\mathbf{B}_{i} \in \{0, 1\}^{W \times L \times H}$ defined in its local coordinate. Here $S$ is the total number of teammates, the dimensions $W$, $L$, and $H$ respectively denote the width, length, and height of the BEV map.

\textbf{Collaborative 3D detection.} A collaborative detector shared amongst vehicles is composed of an encoder denoted by $f_\psi$, an aggregator and a decoder which are collectively denoted by $f_\theta$ for simplicity. $f_\psi$ will take the BEV map as input and generate an intermediate feature map $\mathbf{M}_i\in \mathbb{R}^{\frac{W}{K} \times \frac{L}{K} \times C} = f_\psi(\mathbf{B}_{i})$ as the transmitted message, where $K$ indicates the downsampling scale in the neural network, and $C$ denotes the feature dimension. After deciding to use the messages from $s$ teammates $\{\mathbf{M}_j \}_{j=1,\ldots,s}$ ($s\leq S$), the ego-vehicle indexed by $0$ will use $f_\theta$ to produce a set of bounding boxes $\hat{\mathbf{Y}}_s = f_\theta(\mathbf{M}_0, \{ \mathbf{M}_j \}_{j=1,\ldots,s})$. During training, $f_\psi$ and $f_\theta$ are jointly learned by minimizing the detection loss $\mathcal{L}_{det}(\hat{\mathbf{Y}}_s, \mathbf{Y}_{gt})$, where $\mathbf{Y}_{gt}$ denotes ground-truth boxes. $\mathcal{L}_{det}$ includes a classification loss and a regression loss following existing detectors~\cite{Luo2018FastAF}.

\textbf{Adversarial attack.} We consider the white box attack where the attacker has full access to the target model because the detector model is shared amongst vehicles. The adversarial message is generated by gradient-based optimization  to maximize the vehicle's detection errors. Specifically, at inference, model parameters are frozen, and the objective for attacker $v$ is to fool the ego-vehicle by sending an indistinguishable adversarial message $\mathbf{M}_v + \delta$:
\begin{equation}
    \max_{||\delta|| \leq \Delta} \mathcal{L}_{det}(f_\theta(\mathbf{M}_0, \mathbf{M}_v + \delta, \{ \mathbf{M}_j \}_{j\neq v}), \mathbf{Y}_{gt}) , 
\end{equation}
where $\delta \in \mathbb{R}^{\frac{W}{K} \times \frac{L}{K} \times C}$ with the same size as $\mathbf{M}_v$ is the optimized perturbation, and is constrained by $\Delta$ to ensure its imperceptibility. In practice, the detection results of solely using the message of ego-vehicle $\hat{\mathbf{Y}}_0 = f_\theta(\mathbf{M}_0)$ could replace $\mathbf{Y}_{gt}$ in case the ground-truth is not available. Such kind of carefully-crafted adversarial messages can create a lot of false positives/negatives for the ego-vehicle, which raises concerns for safety-critical autonomous vehicles.

\textbf{Adversarial defense.} To defend against such kind of adversarial messages, and maximize the usage of complementary messages, the ego-vehicle is supposed to intelligently select teammates to ensure that there are no adversarial messages in the exploited messages $\{\mathbf{M}_j \}_{j=1,\ldots,s}$, while maximizing the number of benign collaborators $s$. To this end, we use the \acronym~ framework shown in Algorithm~\ref{alg:ransac}. 
Specifically, the ego-vehicle obtains $S$ messages $\{ \mathbf{M}_i \}_{i=1,\ldots,S}$ from all the teammates, and it samples $s$ messages each time $\{ \mathbf{M}_j \}_{j=1,\ldots,s}$ and then verify the consensus between two detections $\hat{\mathbf{Y}}_s = f_\theta(\mathbf{M}_0, \{ \mathbf{M}_j \}_{j=1,\ldots,s})$ and $\hat{\mathbf{Y}}_0 = f_\theta(\mathbf{M}_0)$. Since the adversarial learning will result in a large number of false detections, $\hat{\mathbf{Y}}_s$ will significantly differ from $\hat{\mathbf{Y}}_0$ once there exist adversarial messages in $\{ \mathbf{M}_j \}_{j=1,\dots,s}$. In 3D detection, the difference measure $d$ is represented by the
% \qi{Not average, any single box overlapped $>$ IoU thresh is considered a match item}  
intersection-over-union (IoU) between two sets of boxes after Hungarian matching~\cite{kuhn1955hungarian}, and a consensus threshold $\epsilon$ is used to determine whether or not there is consensus. The ego-vehicle will keep on sampling until the sampled vehicles reach a consensus. In practice, we may want to give a certain sampling budget $N$ to avoid too many computations for $\hat{\mathbf{Y}}_s = f_\theta(\mathbf{M}_0, \{ \mathbf{M}_j \}_{j=1,\ldots,s})$, and the ego-vehicle can compute the maximum number of benign collaborators which could be found, and use it as the sample size $s$ based on Eq.~\ref{eq:sampling_size},  with the given sampling budget as well as the attacker ratio. In a word, the ego-vehicle can have $s$ collaborators in consensus to aid in the promotion of its perception within $N$ sampling steps.

%===============================================================================

\section{Experiments}
\subsection{Experimental setup}
\textbf{Dataset and detector.} We employ V2X-Sim~\cite{li2022v2x} to verify our method. The dataset contains 5Hz LiDAR point clouds recorded  by different vehicles at the same intersection. Regarding the multi-robot detector, we employ a simple average-based collaborative perception method which calculates the mean of the feature maps from different vehicles and feeds the aggregated features into the decoder to generate the final perception results. The detector backbone is a simple anchor-based method~\cite{Luo2018FastAF}. We follow the training procedures and evaluation protocols in~\cite{li2021learning}.

\textbf{Adversarial attack implementation.} We use three attackers: projected gradient descent (PGD)~\cite{madry2017towards}, Carlini \& Wagner Attack (C\&W)~\cite{carlini2017towards}, and basic iterative method (BIM)~\cite{kurakin2018adversarial}. The number of iterations in PGD/BIM is set to $15$ (iteration of C\&W is set to $30)$, the step size is set to $0.1$, and the magnitude is $\Delta = 0.5$. The consensus threshold is set to $\epsilon = 0.3$. We use average precision (AP) at IoU thresholds of 0.5 and 0.7 to evaluate the detection performance, and we utilize the performance of detection to evaluate the effectiveness of our defense strategies. We adopt scene \#8 with 100 frames (each frame contains 6 vehicles). 
%In each frame, we randomly select one vehicle as the attacker. More experiments are in the supplementary material.

\textbf{Static and dynamic teammates.} Regarding the static team, the ego-robot can establish a stable partnership with reliable collaborators. Once a requisite number of attacker-free teammates is identified, the ego-robot becomes capable of distinguishing potential attackers and subsequently disregards their messages in subsequent frames.  In the scenario of a dynamic team, we need to deploy \acronym~ at each frame, since now the ego-robot fails to rely on its past judgments of the teammates. For the first scenario, the computational burden introduced by \acronym~ is negligible because only several initial frames need consensus verification. Addressing the more complex second scenario requires striking a balance between performance and computational demands under diverse conditions.

\begin{table}[t]
\small
\begin{center}
\resizebox{0.36\textwidth}{!}{%
            \begin{tabular}{@{}ccccccc@{}}
                  \toprule
                  \multicolumn{3}{c}{{\bf ROBOSAC}} &  \multicolumn{3}{c}{{\bf Actual Steps}} &
                  {\bf Success}
                %  {\bf Method}   &    {\bf mAP@0.5}  & {\bf mAP@0.7}
                  \\
                \bf $\eta$ & \bf $s$ & \bf $N$ & \bf Avg & \bf Min & \bf Max & {\bf Rate} \\
                \midrule
                \multirow{4}{*}{0.2}
                & 1 & 3 & 1.32 & 1 & 6 & 0.96 \\
                & 2 & 5 & 1.76 & 1 & 6 & 0.97    \\
                & 3 & 7 & 2.31 & 1 & 7 & 1.00     \\
                & 4 & 9 & 4.89 & 1 & 19 & 0.89   \\
                \midrule
                \multirow{3}{*}{0.4}
                & 1 & 6 &1.80 & 1 & 8 & 0.98 \\
                & 2 & 11&3.06 & 1 & 11 & 1.00    \\
                & 3 & 19&10.36 & 1 & 39 & 0.86    \\
                \midrule
                \multirow{2}{*}{0.6}
                & 1 & 10 & 2.46 & 1 & 8 & 1.00    \\
                & 2 & 27 & 8.29 & 1 & 46 & 0.97    \\
                \midrule 0.8 
                & 1 & 21 & 4.73 & 1 & 17 & 1.00    \\
                \bottomrule
            \end{tabular}
            % }
% }
}
% }
\end{center}
\vspace{-3mm}
            \caption{\textbf{Validation of our derivation} under different attacker ratio $\eta$ and desired number of collaborators $s$. }
            \label{tab:validation}
\end{table}

\begin{table}[t]
\small
%   \fontsize{8}{9.6}
%   \fontsize{4}{4.8}\selectfont
        \begin{center}
        \small
            \resizebox{0.36\textwidth}{!}{%
            {%\renewcommand{\arraystretch}{1.005}
            \setlength{\tabcolsep}{1mm}{
            \begin{tabular}{@{}ccccc@{}} 
            \toprule 
            \multirow{2}{*}{\textbf{Setup}}&\multicolumn{2}{c}{\textbf{AP}}&\textbf{Avg.}&\multirow{2}{*}{\textbf{FPS}}\\
            &\textbf{IoU=0.5}&\textbf{IoU=0.7}& \textbf{Steps}\\
            \midrule
            Dynamic team & 77.5 & 74.9 & 2.73 & 14.2\\
            Static team & 78.8 & 76.5 & 2.30 & 36.8\\
            \bottomrule 
            \end{tabular}
            }
        }
        }
        \end{center}
        \vspace{-3mm}
        \caption{\textbf{Comparison between scenarios of dynamic and static team}: averaged over 10 experiments. Agent 1 in scene \#8, with $\eta = 0.2$, $s = 3$, $N = 7$, $\epsilon=0.3$}
        \label{tab:offline}
        \vspace{-4mm}
\end{table}

\subsection{Quantitative results}\label{sec:result}

\textbf{Validation of the derivation.} Given the attacker ratio $\eta$ and the desired number of collaborators $s$, the maximum number of sampling  $N$ could be given by Eq.~\ref{eq:sampling_number} based on a certain probability $p=0.99$. We validate this formulation by allowing the ego-vehicle to keep on sampling until reaching consensus, under different $\eta$ and $s$, and the results are shown in Table~\ref{tab:validation}. For each processed frame, it is considered a success if the actual steps taken to achieve consensus are not larger than the computed $N$, and the ratio of successful frames is referred to as the \textit{success rate}. We see that a consensus can be mostly achieved within the theoretical upper-bound $N$. Meanwhile, the average sampling steps taken for consensus are usually less than 50\% of $N$. 

\begin{table}[t]   
\small
\centering
            \begin{tabular}{@{}cccc@{}} 
            \toprule 
            \multirow{2}{*}{\textbf{Method}}&\multicolumn{2}{c}{\textbf{AP}}&\textbf{Success}\\
            &\textbf{IoU=0.5}&\textbf{IoU=0.7}& \textbf{Rate}\\
            \midrule
            Upper-bound++&81.3&79.8&$-$\\
            Upper-bound&78.0&76.0&$-$\\
            \midrule
            Sampling budget: 7&77.3&74.7&1.00\\
            Sampling budget: 5&76.6&74.1&0.96\\
            Sampling budget: 3&74.8&73.2&0.77\\
            Sampling budget: 1&69.8&67.4&0.45\\
            \midrule
            Lower-bound&63.5&60.1&$-$\\
            No Defense&39.7&39.0&$-$\\
            \bottomrule
            \end{tabular}
        \caption{\textbf{Detection performance with different sampling budgets} when $\eta=0.2$ and $s=3$. Upper-bound++ denotes collaborative perception with all 5 benign robots. Upper-bound means collaborative perception with 3 out of 5 benign robots. Lower-bound denotes individual perception.}
        \label{tab:tradeoff}
        % \vspace{-4mm}
\end{table}

\begin{figure}[t]
    \begin{center}
        \includegraphics[width=0.45\textwidth]{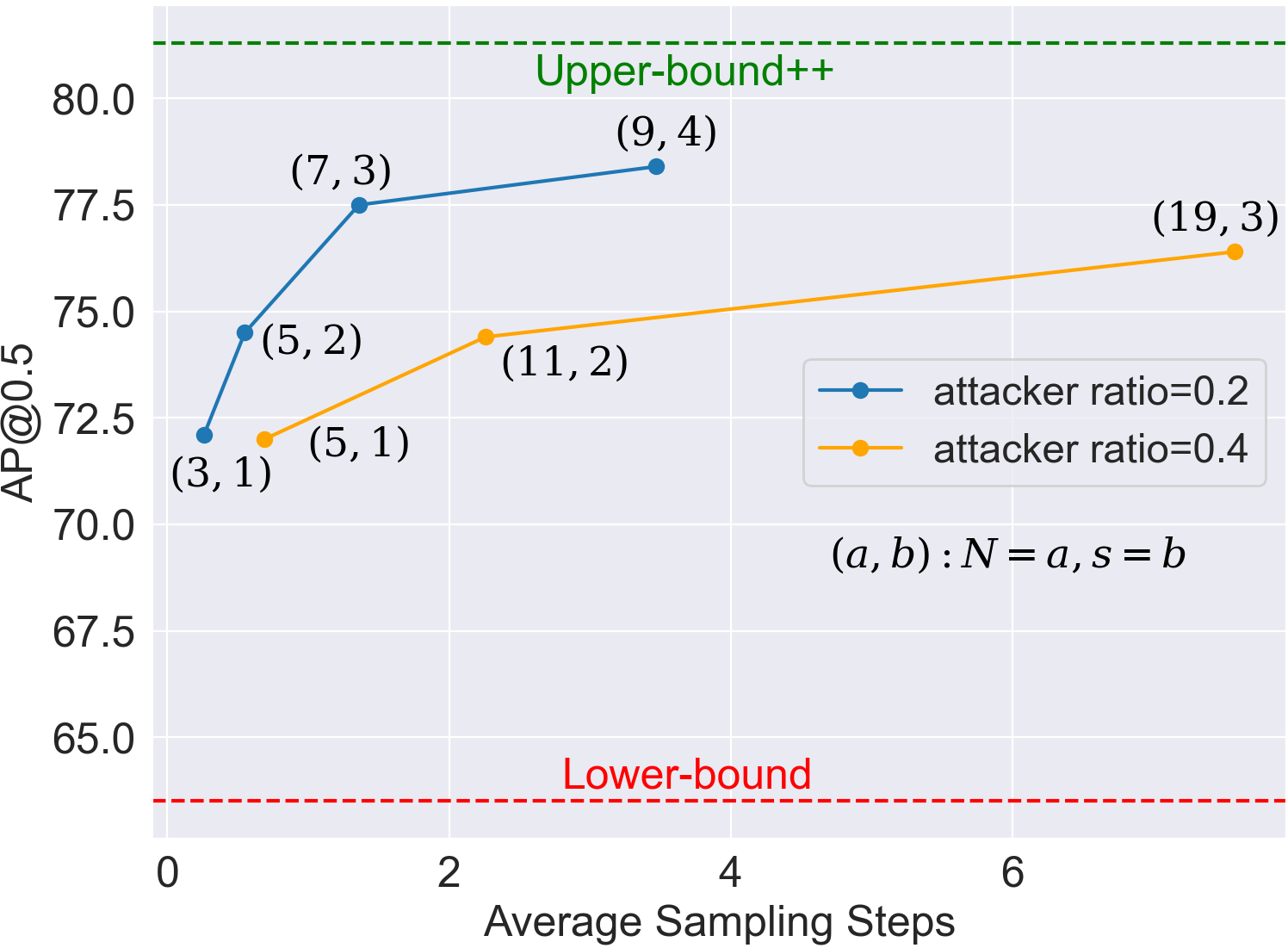}
        \captionof{figure}{\textbf{Performance-computation trade-off plot.}}
        \label{fig:tradeoff}
        \end{center}
        % \caption{Precision-computation trade-off plot.}
        \vspace{-8mm}
\end{figure}

\textbf{Static team.} Since the teammates are not changed, so once the attackers are identified, the ego-robot could avoid being attacked in the following frames. The results  are shown in Table~\ref{tab:offline}. We see that the frames-per-second (FPS) is 36.8, which is satisfactory for real-time applications if the \acronym~ is only needed at the initial stage.

\textbf{Dynamic team}. In the scenario of dynamic teammates, the ego-robot needs to identify the collaborators at each frame. Since the actual sampling steps are usually much less than the upper bound, we can lower the sampling budget in practice to pursue a balance between performance and computation. Assume we aim to find three attacker-free teammates when $\eta = 0.2$, the upper bound of the sampling budget is then $N=7$. To save computations, we set the sampling budget as $N=1,3,5,7$ respectively and report the detection performance as well as the success rate (successfully find three attacker-free teammates within the specified sampling budget) in Table~\ref{tab:tradeoff}. We find that limiting the sampling budget to 5 still maintains a success rate of $0.96$. Meanwhile, only sampling once in each frame (about $15\%$ of $N$) can still achieve a success rate of $0.45$ and an AP of $69.8/67.4$ which are still better than solely using ego-vehicle's information ($63.5/60.1$). In addition, we conduct precision-computation trade-off analyses under different $N-s$ pairs. As shown in Fig.~\ref{fig:tradeoff}, detection precision is higher when there are more collaborators, at the cost of more computations to reject attackers.

\begin{table}[t]
\small
    \centering
    
    \fontsize{10}{12}\selectfont
        % 	\vspace{-2mm}
        \begin{center}
        \resizebox{0.45\textwidth}{!}{%
        {\renewcommand{\arraystretch}{1.005}
            \setlength{\tabcolsep}{1mm}{
            \begin{tabular}{@{}ccccc@{}} 
            \toprule 
            \multirow{2}{*}{\textbf{Method}}&\multicolumn{2}{c}{\textbf{AP}}&\textbf{Success}&\multirow{2}{*}{\textbf{FPS}}\\
            &\textbf{IoU=0.5}&\textbf{IoU=0.7}& \textbf{Rate}\\
            \midrule
            w/ Temporal Consistency & 75.8 & 73.6 & 0.85 & 19.2\\
            w/o Temporal Consistency & 77.5 & 74.9 & 0.98 & 14.2\\
            \bottomrule 
            \end{tabular}
            }
            }
            }
            \end{center}
            \vspace{-3mm}
            \caption{\textbf{Comparison between w/ temporal consistency and w/o temporal consistency}: agent 1 in {scene \#8}, with $\eta = 0.2$, $s = 3$, $N = 7 $, $\epsilon=0.3$}
            \label{tab:temporal}
        	% \vspace{-4mm}
\end{table}

\begin{table}[t]{
\small
\begin{center}
\resizebox{0.45\textwidth}{!}{%
\setlength{\tabcolsep}{1mm}{
\begin{tabular}{@{}ccc@{}} 
            \toprule 
            \multirow{2}{*}{\textbf{Method}}&\multicolumn{2}{c}{\textbf{AP}}\\
            &\textbf{IoU=0.5}&\textbf{IoU=0.7}\\
            \midrule
            Upper-bound++&81.8&79.6\\
            \midrule
            \acronym~ (against PGD attack)&77.9&75.6\\
            PGD Trained (White-box Defense)&75.6&73.0\\
            \midrule
            \acronym~ (against C\&W attack)&74.5&71.1\\
            C\&W on PGD Trained (Black-box Defense)&43.2&40.8\\
            \midrule
            Lower-bound&64.1&62.0\\
            No Defense (PGD attack)&44.2&43.7\\
            \bottomrule
\end{tabular}
}
}
\end{center}
}
\vspace{-3mm}
\caption{\textbf{Quantitative results of generalizability test} on agent 1 in {scene \#8}, with $\eta = 0.2$, $s = 3$, $\epsilon=0.3$. The average sampling step and FPS are 2.67 and 15.5. Black-box defense is unaware of the attacker, while white-box defense knows the attacker and employs the  adversarial training. Upper-bound++ denotes collaborative perception with all 5 benign robots. Lower-bound is individual perception. }
\label{tab:generalization}
\vspace{-5mm}
\end{table}

\begin{table}[t]
\small
    \centering
            \begin{tabular}{@{}cccc@{}} 
            \toprule 
            \multirow{2}{*}{\textbf{Ablation}}&\multicolumn{2}{c}{\textbf{AP}}&\textbf{Success}\\
            &\textbf{IoU=0.5}&\textbf{IoU=0.7}& \textbf{Rate}\\
            \midrule
            Threshold $\epsilon =0.1$ & 77.5 & 75.4 & 0.96\\
            Threshold $\epsilon =0.2$ & 78.0 & 75.0 & 0.98\\
            Threshold $\epsilon =0.3$ & 77.8 & 76.1 & 1.00\\
            Threshold $\epsilon =0.4$ & 76.7 & 74.7 & 0.95\\
            Threshold $\epsilon =0.5$ & 76.1 & 73.9 & 0.87\\
            \midrule
            PGD, 10 iterations & 78.3 & 75.2 & 0.95\\
            PGD, 15 iterations & 77.8 & 76.1 & 1.00\\
            BIM, 10 iterations & 77.8 & 75.2 & 0.95\\
            BIM, 15 iterations & 77.5 & 74.5 & 0.97\\
            \bottomrule 
            \end{tabular}
        \caption{\textbf{Quantitative results of ablation studies.}  }
        \label{tab:ablation}
        % \vspace{-4mm}
\end{table}

\begin{table}[t]{
\begin{center}
\resizebox{0.45\textwidth}{!}{%
\setlength{\tabcolsep}{1mm}{
\begin{tabular}{@{}ccccccc@{}} 
            \toprule 
            \multirow{2}{*}{\textbf{Metrics}}&\multicolumn{5}{c}{\textbf{\ \ \ Attacker Ratio}}\\
            &\textbf{0.0}&\textbf{0.2}&\textbf{0.4}&\textbf{0.6}&\textbf{0.8}&\textbf{1.0}\\
            \midrule
            Frame \# to reach final estimation &0&0.8&2.1&3.1&1.3&0\\
            Final estimated ratio&0.0&0.22&0.42&0.60&0.80&1.0\\
            Error of estimation&0.0&0.02&0.02&0.0&0.0&0.0\\
            Total sampling steps&1&8.2&20.5&37.9&59&77\\
            Success Rate&1.0&0.9&0.9&1.0&1.0&1.0\\
            \bottomrule
\end{tabular}
}
}
\end{center}
}
\vspace{-3mm}
\caption{{\textbf{Quantitative results of attacker ratio estimation} on agent 1 in {scene \#8}, with $N = 5, \epsilon=0.3, \mathbf{R} = \left[0.0, 0.2, 0.4, 0.6, 0.8\right]$. All results are averaged over 10 repeated experiments.} }
\label{tab:probing}
\vspace{-5mm}
\end{table}

\textbf{Temporal consistency.} We further propose to use temporal consistency instead of the difference between collaborative and individual perception to save computations. Specifically, we compare the current output with the previous output for consensus verification. This can help to improve efficiency because we can reduce the times of model forward caused by individual perception. The results are shown in Table~\ref{tab:temporal}. We see that FPS is improved from 14.2 to 19.2, yet the performance is still comparable to using individual perception results as a reference.

\textbf{Generalizability.} We compare the generalizability of \acronym~ to adversarial training. We use PGD~\cite{madry2017towards} which is the strongest one-stage gradient-based adversarial attack. The results are shown in Table~\ref{tab:generalization}: although adversarial training with PGD could effectively defend the PGD attack with a precision 75.6 (IoU=0.5), using a different attacker, Carlini \& Wagner Attack~\cite{carlini2017towards}, will largely degrade the precision to 43.2 (IoU=0.5). In contrast, our method can achieve comparable precision under both two attackers (77.9@IoU0.5 under PGD attack, and 74.5@IoU=0.5 under C\&W attack). Our better generalizability stems from being attacker-agnostic: we do not rely on the knowledge of attackers while adversarial training does.

% \Note{sc: need to definer upperbound and lowerbound somewhere in the text.}

\textbf{Attacker ratio estimation.} In practice, using a budget of five samplings within a single frame yields acceptable results. As shown in Table~\ref{tab:probing}, within the initial few frames, the actual attacker ratio can be efficiently probed. The estimated ratio can then be used to carry out \acronym~ steps.

\begin{figure*}[t]
\scriptsize
\centering
   \vspace{-2mm}
\includegraphics[width=0.95\textwidth]{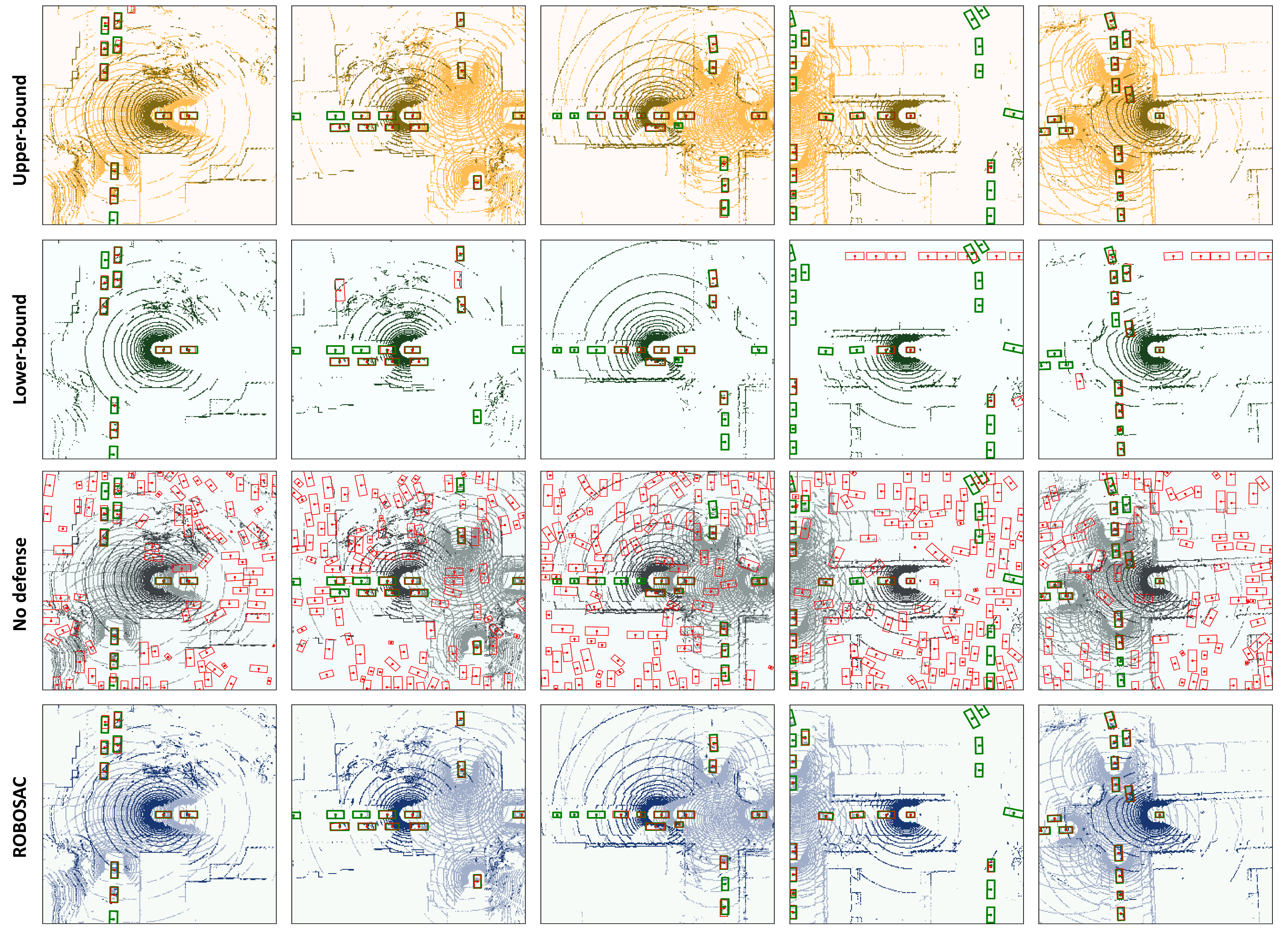}
	\vspace{-2mm}
\caption{\textbf{Visualization of perception results} on V2X-Sim~\cite{li2022v2x}. Red boxes denote predictions, and green boxes are ground truth. Upper-bound, lower-bound have the same meanings with that in Table~\ref{tab:tradeoff}.}
\label{fig:qualitative}
	\vspace{-3mm}
\end{figure*}

\subsection{Ablation studies}
We conduct ablation studies on consensus threshold as well as attack approaches (the other parameters are $N = 7$, $s = 3$, $\eta = 0.2$), and the results are shown in Table~\ref{tab:ablation}. Regarding the consensus threshold, a lower threshold indicates a lesser tolerance for outcomes that differ from individual results, whereas a higher threshold indicates a higher likelihood of believing the variation is good. We find that our method achieves comparable performance with different consensus threshold. Meanwhile, our method remains unaffected by the type of attacker. 

\subsection{Computational Cost}
Our method's performance depends on the sampling steps, each requiring a forward propagation. Using an NVIDIA RTX 3090 GPU, the baseline detector averages 17ms for ego-only predictions and 27ms for collaborative predictions. With a 5Hz dataset, we allocate a 200ms time budget per frame, allowing up to 7 samplings. As actual sampling steps are lower, we can achieve a better performance-computation trade-off by sampling more attacker-free teammates, as shown in Fig.~\ref{fig:tradeoff}.

% \Note{sc: compare the computational cost with adversarial training? }

\section{Limitations}
We assume that the attacker ratio is fixed, yet the attacker ratio may vary in practice. Besides, we assume that although the input adversarial noise is imperceptible, its effect on the network output is significant (see Fig.~\ref{fig:qualitative}), which has been observed in most existing attacking methods. Future attackers might develop dangerous yet subtle perturbations in both the input and output to bypass our ``outlier-detection-based'' defense mechanism, although currently, we are not aware of any such attacks.
% Future works may consider more advanced attackers which results in subtle yet dangerous change in output space.

%===============================================================================

\section{Conclusion}\label{sec:conclusion}
In this work, we propose a novel adversarially robust collaborative perception framework termed \acronym~. It makes as much use of messages from benign collaborators as possible while resisting adversarial attackers within a certain computation budget. Moreover, we develop an aggressive-to-conservative probing method with retrospect for attacker ratio estimation in scenarios of unknown ratios. We validate our method in collaborative 3D detection for autonomous driving. Unlike adversarial training, our approach relies on the consistency in output space rather than the knowledge of a specific adversarial noise, thus is more generalizable. We believe our work will further improve the adversarial robustness of multi-robot systems.

\balance
{\small
\bibliographystyle{unsrt}
\bibliography{egbib}
}

\end{document}

%% file: packages.tex
\usepackage{iccv}
\usepackage[accsupp]{axessibility} 
\usepackage{graphics} % for pdf, bitmapped graphics files
\usepackage{epsfig} % for postscript graphics files
\usepackage{empheq}
\usepackage{times} % assumes new font selection scheme installed
\usepackage{amsmath} % assumes amsmath package installed
\usepackage{amssymb}  % assumes amsmath package installed
\usepackage{bm}
\usepackage{color}
\usepackage[table,dvipsnames]{xcolor}
\usepackage{bbding}
\usepackage{adjustbox}
\usepackage{changepage}
\usepackage{cite}
\usepackage{diagbox}
\usepackage{enumitem}
\usepackage{hyperref}
\hypersetup{
colorlinks=true,
linkcolor=blue,
filecolor=blue,      
urlcolor=NavyBlue,
citecolor=[RGB]{119,185,0},
}
\usepackage{cleveref}
\usepackage{sidecap}
\usepackage{float}
\usepackage{booktabs}
\usepackage{multirow}
\usepackage{ bbold }
\usepackage{mathrsfs}
\usepackage[utf8]{inputenc}
\usepackage{subfigure}
\usepackage{graphicx}
\usepackage{pifont}
\usepackage{threeparttable}
\usepackage{caption}
\usepackage{setspace}
\newcounter{RNum}
\usepackage{balance}
\usepackage{multicol}
\usepackage{blindtext}
\usepackage[linesnumbered,ruled]{algorithm2e}
\usepackage{algorithmic}
\usepackage{amsfonts}
\usepackage{xcolor}
\usepackage{tikz}

\newcommand{\acronym}[1]{ROBOSAC}

\usepackage{etoolbox}  % patch def of algorithmic environment
\makeatletter
\patchcmd{\algorithmic}{\addtolength{\ALC@tlm}{\leftmargin} }{\addtolength{\ALC@tlm}{\leftmargin}}{}{}
\makeatother